\documentclass{article} 
\usepackage{iclr2026_conference, times}

\usepackage{amsmath,amsfonts,bm}

\def\eqref#1{equation~\ref{#1}}

\def\1{\bm{1}}

\DeclareMathAlphabet{\mathsfit}{\encodingdefault}{\sfdefault}{m}{sl}
\SetMathAlphabet{\mathsfit}{bold}{\encodingdefault}{\sfdefault}{bx}{n}

\usepackage{enumitem}
\usepackage{newfloat}
\usepackage{wrapfig}
\usepackage{listings}
\usepackage{xcolor}
\usepackage{colortbl}
\usepackage{microtype}
\usepackage{cleveref}
\usepackage{url}
\usepackage{multirow}
\usepackage{makecell}
\usepackage{todonotes}
\usepackage{tablefootnote}
\usepackage{booktabs}
\usepackage{adjustbox}
\usepackage{enumitem}
\usepackage{arydshln}
\usepackage{array}
\usepackage{ragged2e}
\usepackage{xcolor}
\usepackage{booktabs}
\usepackage{comment}
\usepackage{listings}
\usepackage{inconsolata}
\usepackage{pifont}
\usepackage{amsmath}
\usepackage{wrapfig}
\usepackage{colortbl}
\usepackage{makecell}
\usepackage{amssymb}
\usepackage{graphicx}
\usepackage{booktabs}
\usepackage{lipsum}
\usepackage{multirow}
\usepackage{shadowtext}
\usepackage{xspace}
\usepackage{listings}
\usepackage{fontawesome5}
\definecolor{darkblue}{rgb}{0, 0, 0.5}
\usepackage{afterpage}
\usepackage{graphicx}
\usepackage{algorithm}
\usepackage{algpseudocode}
\usepackage{amsmath}
\usepackage{amsthm}
\usepackage{zref-xr,zref-user}
\usepackage{amssymb}
\usepackage{mathtools}
\usepackage[most]{tcolorbox}

\usepackage{minted}

\setminted{
  breaklines=true,
  frame=lines,
  framesep=2mm,
  fontsize=\small
}

\usepackage{tikz}

\definecolor{beaublue}{HTML}{BCD4E6} 

\title{EvolProver: Advancing Automated theorem proving by Evolving Formalized Problems via Symmetry and Difficulty}

\author{Yuchen Tian$^{2,1}$\thanks{\quad Equal contribution.}  \hspace{0.28em}\thanks{\quad The idea of this work was proposed when  the first author was an intern at Ant Group before joining HKBU, and the work was completed after he joined HKBU.} \quad \bf{Ruiyuan Huang}$^{3, 2}$\footnotemark[1] \hspace{0.28em} \thanks{\quad Work done during internship at Ant Group.}\quad \bf{Xuanwu Wang}$^1$ \quad \bf{Jing Ma}$^1$  \\
\bf{Zengfeng Huang}$^{3,4}$ 
\quad \bf{Ziyang Luo}$^1$ \quad \bf{Hongzhan Lin}$^1$ \quad \bf{Da Zheng}$^2$\footnotemark[4] \quad  \bf{Lun Du}$^2$\thanks{\quad  Corresponding authors. Correspondence to: Da Zheng \textless zhengda.zheng@antgroup.com\textgreater, Lun Du 
\textless dulun.dl@antgroup.com\textgreater
 }\\ 
      $^1$Hong Kong Baptist University   \\
      $^2$Ant Group \\
      $^3$School of Data Science, Fudan University   \\
      $^4$Shanghai Innovation Institute \\
}      

\iclrfinalcopy 
\begin{document}

\maketitle

\begin{abstract}
Large Language Models (LLMs) for formal theorem proving have shown significant promise, yet they often lack generalizability and are fragile to even minor transformations of problem statements. To address this limitation, we introduce a novel data augmentation pipeline designed to enhance model robustness from two perspectives: symmetry and difficulty. From the symmetry perspective, we propose two complementary methods: \textbf{EvolAST}, an Abstract Syntax Tree (AST) based approach that targets syntactic symmetry to generate semantically equivalent problem variants, and \textbf{EvolDomain}, which leverages LLMs to address semantic symmetry by translating theorems across mathematical domains. From the difficulty perspective, we propose \textbf{EvolDifficulty}, which uses carefully designed evolutionary instructions to guide LLMs in generating new theorems with a wider range of difficulty. We then use the evolved data to train \textbf{EvolProver}, a 7B-parameter non-reasoning theorem prover. EvolProver establishes a new state-of-the-art (SOTA) on FormalMATH-Lite with a 53.8\% pass@32 rate, surpassing all models of comparable size, including reasoning-based models. It also sets new SOTA records for non-reasoning models on MiniF2F-Test (69.8\% pass@32), Ineq-Comp-Seed (52.2\% pass@32), and Ineq-Comp-Transformed (34.0\% pass@32). Ablation studies further confirm our data augmentation pipeline's effectiveness across multiple benchmarks.

\end{abstract}

\section{Introduction}
Large Language Models (LLMs) have demonstrated significant potential in mathematical reasoning, sparking a surge of research into their application for formal theorem proving. Formal languages like Lean~\citep{10.1007/978-3-030-79876-5_37}, Coq~\citep{CoqManual6.1}, and Isabelle~\citep{Paulson1994} represent mathematical proofs as rigorous code implementations. This process demands strict syntactic precision and logical soundness, with every proof requiring compiler verification. While this guarantees the absolute reliability of proofs, it also creates a major bottleneck: the extreme scarcity of high-quality training data. Crafting formal proofs requires deep domain expertise and substantial time, a reality that fundamentally conflicts with the data-intensive paradigm of LLMs.


To address the scarcity for data, the research community has explored various data synthesis methods. For instance, DeepSeek-Prover \citep{xin2024deepseek_v1} attempts to automatically translate a large number of informal natural language problems into formal statements, using model scoring and a hypothesis rejection mechanism for screening. Goedel-Prover-V2 \citep{lin2025goedel} adopts a scaffolded strategy to generate mathematical problems of appropriate difficulty to provide models with more effective learning signals. Meanwhile, STP \citep{dong2025stp} constructs two adversarial roles of a conjecturer and a prover that iteratively improve to jointly generate new problems and proofs.

However, a line of work has shown that models trained with such synthesized data still lack generalizability.  For example, \citet{zhao2025ineq} noted that minor transformations of a problem, such as transforming an inequality of the form $f(x) > g(x)$ to $f(x) + f(y) > g(x) + g(y)$, degrade the performance of LLMs drastically. Furthermore, other studies~\citep{informal_robust_1, informal_robust_2} have revealed that this fragility is not unique to formal reasoning; informal LLMs are also susceptible to minor problem transformations.
Motivated by this, we propose a novel data augmentation pipeline to improve model generalizability by addressing it from two perspectives: symmetry and difficulty.

In mathematics, symmetry means exactly invariance under certain transformations. From the symmetry perspective, the fragility of existing models against minor transformations of problems suggests they fail to learn the underlying symmetry structure of the mathematical problem.
To address this, we introduce two complementary methods targeting syntactic and semantic symmetry. 
The first, \textbf{EvolAST}, addresses syntactic symmetry using Abstract Syntax Tree (AST). It parses a formal statement into an AST, applies equivalence transformations using a library of axioms and theorems, and converts the modified tree back into a new statement. This generates semantically identical but syntactically diverse problems. The core strength of EvolAST is its extensibility, as any mathematical equivalence can be encoded as a new transformation rule, allowing for systematic enrichment of the data's structural diversity.

Our second method, \textbf{EvolDomain}, addresses semantic symmetry, where a theorem can be reinterpreted in different domains while preserving its core logic. EvolDomain uses evolutionary instructions to guide LLMs in translating theorems across mathematical domains, thereby creating novel and diverse problem statements.


\begin{figure*}[bt]
    \centering
    \includegraphics[width=\textwidth]{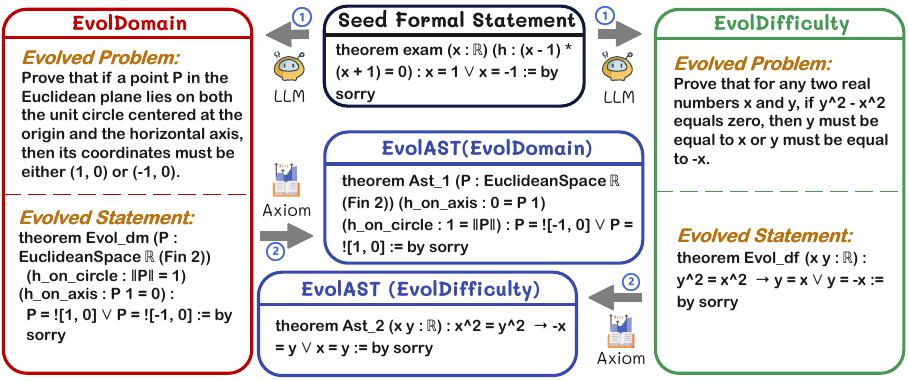}
    \caption{An example of problems evolved by EvolDomain, EvolDifficulty, and EvolAST. A seed formal statement is evolved in parallel by EvolDomain and EvolDifficulty, yielding two new statements. Each of these is then further evolved by EvolAST to generate syntactic variants.}
    \label{fig:evol_example}
    \vspace{-1.5em}
\end{figure*}

From the difficulty perspective, studies have shown that models trained on data with a narrow difficulty range often fail to generalize~\citep{jiang2023_domain_generalization_difficulty_balancing, parashar2025_curriculumreinforcementlearningeasy}. To mitigate this, we propose \textbf{EvolDifficulty}, a method that uses carefully designed instructions to evolve existing theorems by adjusting their difficulty. This process creates a dataset with a much broader difficulty spectrum, which discourages models from relying on shortcuts or mere memorization.



Combining EvolAST, EvolDomain, and EvolDifficulty, we create a comprehensive data augmentation pipeline. Example problems evolved by our pipeline are provided in \Cref{fig:evol_example}. We apply this pipeline to augment public datasets such as STP~\citep{dong2025stp} and Deepseek-Prover-V1~\citep{xin2024deepseek_v1}. By training DeepSeek-Prover-V1.5-Base on this augmented data, we produce our model, \textbf{EvolProver}.
EvolProver achieves state-of-the-art (SOTA) performance on multiple benchmarks. Notably, EvolProver is a non-reasoning (i.e., non-CoT) model, yet it achieves results comparable to, and sometimes surpassing, those of reasoning models. On FormalMATH-Lite~\citep{yu2025formalmath}, it sets a new SOTA with a 53.8\% pass@32 rate among
models of comparable size, including reasoning models. Furthermore, it establishes new SOTA pass@32 rates for non-reasoning models of comparable size on several benchmarks: 69.8\% on MiniF2F-Test~\citep{zheng2021minif2f}, 52.2\% on Ineq-Comp-Seed~\citep{zhao2025ineq}, and 34.0\% on Ineq-Comp-Transformed~\citep{zhao2025ineq}. Ablation studies confirm the efficacy of our pipeline, showing that EvolProver outperforms its counterparts trained on unaugmented or partially augmented data, in some cases by over 10 percentage points.

The main contributions of this work can be summarized as follows:
\begin{itemize}

\item We propose a novel data augmentation pipeline that improves model generalizability by systematically enhancing formalized data directly from both symmetry and difficulty perspectives.

\item We propose EvolAST, a highly extensible, AST-based method that generates syntactically diverse yet semantically equivalent problems by leveraging formal axioms and theorems as transformation rules. Additionally, we introduce EvolDomain and EvolDifficulty, two LLM-driven methods that enrich training data by translating problems across domains and evolving their difficulty, respectively.

\item We train and release EvolProver, a powerful non-reasoning theorem prover built on our augmented data. EvolProver achieves state-of-the-art performance across multiple benchmarks, outperforming all comparable models on FormalMATH-Lite and setting new records for non-reasoning models on others.

\end{itemize}

\section{Related Works}
 
\paragraph{Formal Provers.} Numerous LLM-based formal provers\citep{ji2025leanabell,zhang2025leanabell,shang2025stepfun} have emerged after the advent of ChatGPT, including reasoning-based models like DeepSeek-Prover-V2~\citep{ren2025deepseek}, non-reasoning models like STP~\citep{dong2025stp}, and tree-search models like BFS-Prover~\citep{xin2025_BFS_Prover}. Our work focuses on advancing the state-of-the-art for non-reasoning models, which offer significant computational efficiency.

\textbf{Data Augmentation in Mathematical Reasoning.}
The critical need for large-scale, high-quality training data has spurred significant research into automated methods for mathematical problem generation. Prominent approaches in informal mathematics include MetaMath~\citep{yu_2024_metamath}, which bootstraps new data by rewriting existing questions from multiple perspectives like rephrasing and backward reasoning. Similarly, WizardMath~\citep{luo_2025_wizardmath} adapts the Evol-Instruct framework~\citep{WizardCoder,WizardLM} to systematically generate problems of varying complexity. Another work, PromptCoT~\citep{zhao_2025_promptCoT}, focuses on synthesizing complex problems by emulating the design process of human experts, grounding the generation in core mathematical concepts and logical structures. Inspired by these methods, we introduce EvolDomain and EvolDifficulty. These methods also utilize LLMs but specifically focus on the evolution of formal mathematical statements to enhance their complexity and domain coverage, thereby increasing the diversity of the training data.

While these approaches expand the range and depth of generated problems, they also expose an inherent weakness of LLM-based evolution: the inevitable introduction of syntactic or semantic errors. To mitigate this issue, we propose EvolAST. EvolAST leverages the programmatic features of the Lean 4 proof assistant to perform rewrites directly at the Abstract Syntax Tree (AST) level. This approach ensures that all generated formal statements are syntactically correct and semantically equivalent, effectively increasing data diversity and precision.

\paragraph{Robustness of LLMs in Mathematical Reasoning.} Recent work has highlighted that LLMs lack robustness against small perturbations in mathematical problems, such as variable renaming or adding noise. For instance, the PutnamGAP benchmark~\citep{informal_robust_1} tests equivalence-preserving variants and shows average accuracy declines of 3-10\%. Similarly, MATH-P-Hard~\citep{informal_robust_2} introduces structural shifts, causing performance drops of 10-25\% in models like o1-mini.

While this issue is recognized in informal mathematics, the robustness of LLMs in formal reasoning systems like Lean4 and Coq remains largely underexplored. The Ineq-Comp benchmark~\citep{zhao2025ineq} was developed to address this gap by measuring a prover's performance drop between original problems and their perturbed counterparts.

\section{Method}
\label{method}

\begin{figure*}[bt]
    \centering
    \vspace{-1em}
    \includegraphics[width=\textwidth]{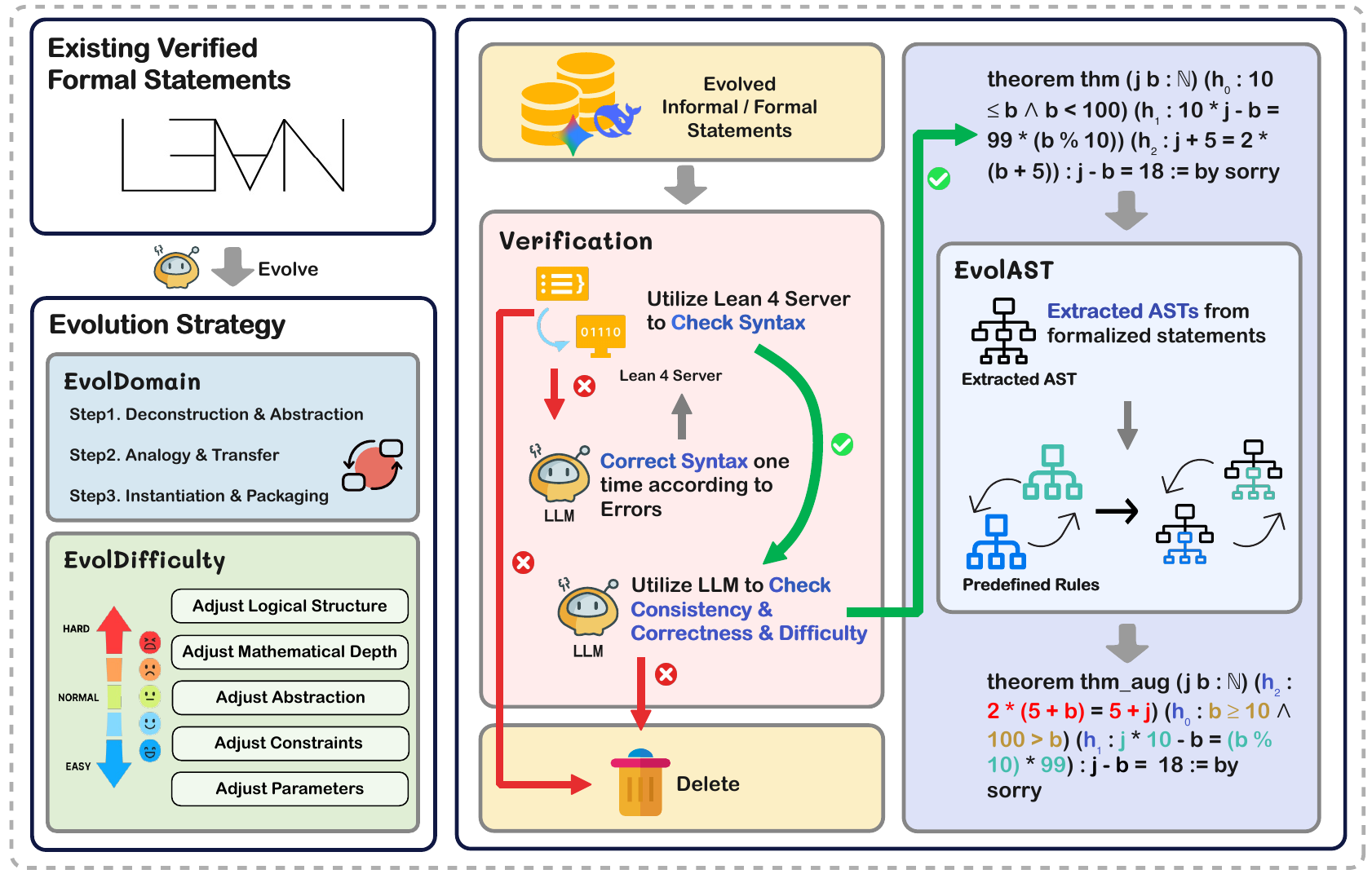}
    \vspace{-1.5em}
    \caption{The workflow of our data augmentation pipeline comprises three phases: EvolDomain and EvolDifficulty, Verification, and EvolAST.}
    \label{fig:pipelin}
    \vspace{-0.5em}
\end{figure*}

Our methodology is centered around a multi-stage data augmentation pipeline, as illustrated in \Cref{fig:pipelin}. First, we leverage LLMs to expand existing formal statements through two evolutionary processes: \textbf{EvolDomain}, for cross-domain translation, and \textbf{EvolDifficulty}, for complexity adjustment. After a rigorous verification stage, we further diversify the data's syntactic structure using \textbf{EvolAST}, a deterministic AST-based transformation method. Finally, we train our model, \textbf{EvolProver}, on this augmented dataset. The following sections detail each component of this pipeline.

\subsection{EvolDomain and EvolDifficulty}
Proven formal statements, with their inherent semantic and syntactic correctness, serve as ideal seeds for data generation. Our work mainly builds upon two open-source datasets, Deepseek-Prover-V1 and STP-Lean, which often lack natural language descriptions. We therefore evolve these formal statements directly by instructing an LLM to generate new, related theorems. 
This approach leverages the logical structure embedded in the formal language itself as a basis for creative generation, bypassing the need for natural language intermediaries.



\paragraph{EvolDomain.}

EvolDomain leverages an LLM to translate a formal statement into new mathematical domains. This process involves three main steps: 1) abstracting the statement's logical skeleton, 2) identifying an analogous concept in a target domain, and 3) instantiating a new, concrete proposition based on this analogy.


Formally, let this process be a function $\mathcal{F}$. Given a source statement $S_i^{\text{formal}}$ and a target domain $D_m$ (selected from a predefined list $\mathcal{L}_D  = \{D_1, D_2, \dots, D_M\}$), $\mathcal{F}$ guides an LLM to first extract the statement's abstract logical skeleton. Based on this skeleton, the model identifies a structurally similar concept in $D_m$ and uses it to construct a new proposition. The output is a pair consisting of a natural language description, $\widehat{P}_i$, and a new formal statement, $\widehat{S}_i^{\text{formal}}$. This can be formally represented as: $\mathcal{F}(S_i^{\text{formal}}, D_m) = (\widehat{S}_i^{\text{formal}}, \widehat{P}_i).$

To maximize the exploration of logical connections across domains, our prompt further guides the LLM to simultaneously transfer and instantiate the core logical skeleton into 3 to 5 distinct new domains. Therefore, the final output of a single function call is a set of pairs spanning multiple domains, with each pair containing a new formal statement and its corresponding natural language description. Prompt templates and examples can be found in Appendix~\ref{appendix:details_for_evoldomain}.

\paragraph{EvolDifficulty.} EvolDifficulty leverages an LLM to adjust a formal statement's difficulty, thereby creating a dataset with a broad difficulty spectrum. We denote this process by the function $\mathcal{E}$.  The process,  $\mathcal{E}$, is guided by carefully designed evolution strategies. Based on expert consultation, we designed five core evolution strategies, $\mathcal{S} = \{s_1, \dots, s_5\}$: (1) Adjusting Logical Structure, (2) Adjusting Mathematical Depth, (3) Adjusting Abstraction, (4) Adjusting Constraints, and (5) Adjusting Parameters. Given a formal statement $S_i^{\text{formal}}$, the function applies a strategy $s_k \in \mathcal{S}$ with an evolution direction $\delta \in \{+1, -1\}$ (for increasing or decreasing difficulty, respectively) to instruct an LLM to generate a new pair of a new formal statement $\widehat{S}_i^{\text{formal}}$ and its natural language description $\widehat{P}_i$.  This can be formally represented as $\mathcal{E}(S_i^{\text{formal}}, s_k, \delta) = (\widehat{S}_i^{\text{formal}}, \widehat{P}_i).$

By systematically applying this framework, EvolDifficulty enables fine-grained control over dataset difficulty, generating problems with a smooth gradient that enriches the dataset's hierarchical structure. Prompt templates and examples can be found in Appendix~\ref{appendix:details_for_evoldifficulty}.

\textbf{Verification.} We employ a stringent two-stage verification pipeline to ensure data quality. First, each generated statement $\widehat{S}i^{\text{formal}}$ is validated for syntactic integrity using the Lean 4 compiler. Statements that fail are given a single LLM-based repair attempt before being discarded. Second, all syntactically valid pairs $(\widehat{S}_i^{\text{formal}}, \widehat{P}_i)$ undergo semantic evaluation by an LLM-based judge. The judge assesses three aspects: consistency between the formal and natural language versions, propositional correctness, and difficulty appropriateness. This dual-filter mechanism, combining deterministic compilation with semantic judgment, ensures that only syntactically sound and semantically coherent data populates our final dataset. Prompt templates can be found in Appendix \ref{appendix:details_for_verification}.



\subsection{EvolAST}
EvolAST is founded on the principle that formal language statements, as structured code, can be parsed into Abstract Syntax Tree (AST). This allows us to bypass non-deterministic models and instead apply a deterministic set of rewriting rules based on established axioms and theorems, guaranteeing semantic equivalence.

We formalize this process as a function $\mathcal{A}$. EvolAST implements an extensible set of rewriting rules (currently 7 rules), $\mathcal{R} = \{r_1, \dots, r_7\}$, where each rule $r_k$ corresponds to a specific logical equivalence: (1) Hypothesis Reordering, (2) Commutativity, (3) Associativity, (4) Distributivity, (5) De Morgan's Laws, (6) Operand Swapping for Symmetric Relations, and (7) Dual Relation Conversion. Given an input statement $S_i^{\text{formal}}$, the function $\mathcal{A}$ first parses it into an AST. It then recursively traverses the tree, applying any applicable rule $r_k \in \mathcal{R}$ at each node with a predefined probability $p$. Finally, the modified AST is recompiled into a new formal statement $\widehat{S}_i^{\text{formal}}$. The process can be formally represented as $
\mathcal{A}(S_i^{\text{formal}}, p) = \widehat{S}_i^{\text{formal}}.$ We provide an example in Appendix~\ref{appendix:details_for_AST}.

Since all transformations are based on strict logical equivalences, EvolAST generates syntactically diverse data while ensuring semantic correctness, thus eliminating the need for further verification. The framework is highly extensible, as any known mathematical or logical equivalence can be encoded as a new rewriting rule.

\subsection{Training EvolProver}
We trained our final model, EvolProver, by fine-tuning DeepSeekProver-V1.5-Base~\citep{xin2024deepseek_v1_dot_5} on our augmented dataset. The training process consists of two stages: Supervised Fine-Tuning (SFT) and Reinforcement Learning (RL). Detailed information on dataset curation and training algorithms can be found in Appendix~\ref{subsec:training_evolprover}.

For comparison and ablation studies, we also trained several other models. This includes a baseline model, \textbf{EvolProver-Base}, which was trained exclusively on the original, unaugmented public data. We also prepared a series of specialized models for our comprehensive ablation experiments, with details provided in Appendix~\ref{subsec:Ablation_Experiments}.




\section{Experiments}
\label{experiments}



\subsection{Baselines}
Existing formal provers are broadly categorized into three types: non-reasoning, reasoning, and tree-search models.

\textbf{Non-reasoning models} generate proofs end-to-end without an intermediate thought process. Key examples include DeepSeek-Prover-V2 (non-CoT)~\citep{ren2025deepseek}, Goedel-Prover-SFT~\citep{lin2025goedel}, and STP~\citep{dong2025stp}.

\textbf{Reasoning models} employ a chain-of-thought process to generate proofs, where the reasoning process is often significantly longer than the final proof. Key examples are DeepSeek-Prover-V2 (CoT)~\citep{ren2025deepseek}, Moonshot's Kimi-Prover-Preview~\citep{kimina_prover_preview_2025} and Kimi-Prover~\citep{kimina_prover_2025}, and Goedel-Prover-V2~\citep{lin2025goedel}. Notably, DeepSeek-Prover-V2 has both a reasoning and a non-reasoning mode.
While generally higher performing, reasoning models demand substantial computational resources due to their chain-of-thought approach (e.g., more than 6000 tokens per proof vs. less than 700 for non-reasoning models). This focus on token efficiency has spurred a recent wave of interest in fast, non-reasoning models, such as Claude 4's Non-thinking mode~\citep{anthropic_claude4_2025} and Grok-Code-Fast-1~\citep{grok_code_fast}.

\textbf{Tree-search models} represent an intermediate proof state as a node in a search tree and use a model to assign heuristic scores to guide the search order. Key examples include BFS-Prover~\citep{xin2025_BFS_Prover}, DeepSeek-Prover-V1.5 + RMaxTS~\citep{xin2024deepseek_v1_dot_5}, and InternLM2.5-StepProver~\citep{Wu2024_InterLM2.5_StepProver}.

For our comparative analysis, we report the performance metrics as published by the original authors to ensure consistency and avoid discrepancies from our own re-evaluations.


\subsection{Results}
\label{subsec:results}
\textbf{FormalMATH}~\citep{yu2025formalmath} is a broad dataset of formal theorems. We follow standard practice and evaluate on its 425-problem subset, FormalMATH-Lite, as other problems in the full dataset were used in training. Problems within FormalMATH-Lite were held out and used exclusively for final evaluation.


The results are summarized in \Cref{table:formalmath_result}. EvolProver achieves a new SOTA of 53.86\% among models of comparable size, surpassing the previous best of 51.76\%. Notably, our non-reasoning model outperforms top reasoning models like DeepSeek-Prover-V2 and Kimi-Prover-Preview. Furthermore, EvolProver outperforms its baseline, EvolProver-Base, by 9.14 percentage points, demonstrating the significant impact of our data augmentation pipeline.

\begin{table}
    \centering
    \caption{Comparison with SOTA 7B-size models on the FormalMATH-Lite dataset; \textcolor{red!85!black}{$\uparrow$} means increase in absolute performance over the ablation model EvolProver-Base; Average Token Length means the average number of output tokens across the benchmark. We do not report average token length for tree-search models, as this metric is not directly comparable with other model types.}
    \label{table:formalmath_result}
    \resizebox{0.9\textwidth}{!}{%
        \begin{tabular}{lccc}
        \addlinespace
        \toprule
        \textbf{Models} & \textbf{Average Token Length} & \textbf{Sample Budget} & \textbf{FormalMATH} \\
        \midrule
        \multicolumn{4}{l}{\textit{Reasoning Models}} \\ 

        ~\makecell[l]{DeepSeek-Prover-V2(COT)}  & 4804.6 & 32 & 51.76\% \\
        ~\makecell[l]{Kimina-Prover-Preview}        &6097.7 & 32 & 48.94\% \\
        \midrule

        \multicolumn{4}{l}{\textit{Tree-Search Models}} \\ 

        ~\makecell[l]{InternLM2.5-StepProver}    & N/A & 1 $\times$ 3200 & 7.87\% \\

        ~\makecell[l]{BFS-Prover}   & N/A & 1 $\times$ 3200 & 27.19\% \\

        \midrule

        \multicolumn{4}{l}{\textit{Non-Reasoning Models}} \\ 
        ~\makecell[l]{DeepSeek-Prover-V1.5-SFT}   &115.9 &32 & 40.40\% \\
        ~\makecell[l]{DeepSeek-Prover-V1.5-RL}    &163.4 &32 & 47.98\% \\
        ~\makecell[l]{Goedel-Prover-SFT}         & 458.4 &32 & 46.70\% \\
        ~\makecell[l]{STP}                       &186.8 &32 & 48.59\% \\

        ~\makecell[l]{\textbf{EvolProver-Base(Ours)}}           &629.8 & 32 & \textbf{44.71\%} \\

        ~\makecell[l]{\textbf{EvolProver(Ours)}}               &653.7 &32 & \textbf{53.86\%}\textcolor{red!85!black}{\scriptsize{($\uparrow$ 9.14\%)}} \\
                                  
        \bottomrule
        \addlinespace

        \end{tabular}
    }
    \vspace{-1em}
\end{table}

\begin{figure*}[t]
    \centering
    \includegraphics[width=\textwidth]{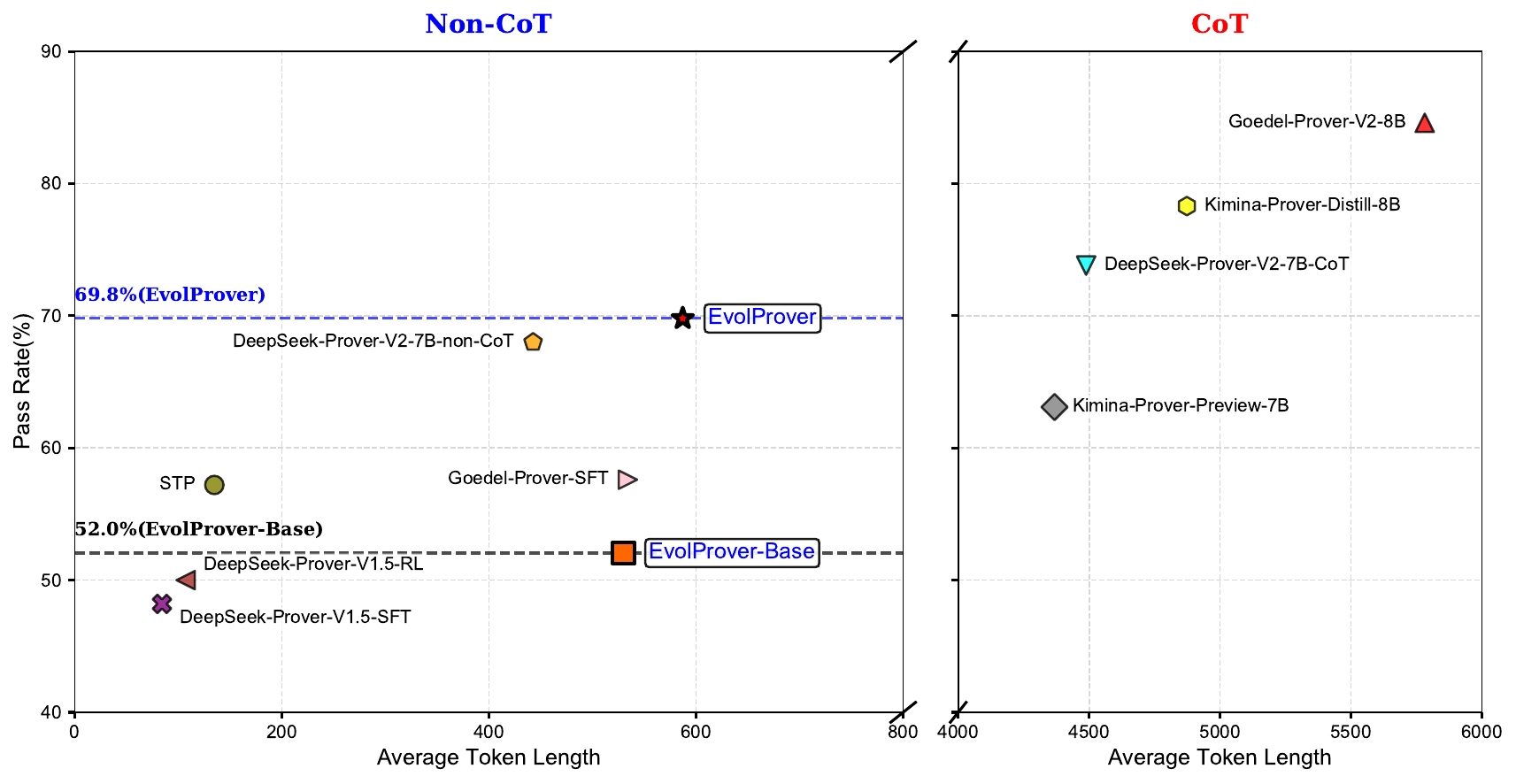}
    \vspace{-2em}
    \caption{Comparison with SOTA models on the MiniF2F-Test dataset. Pass Rate means pass@32 success rate. Average token length is the average number of tokens generated by models across the benchmark. We categorize models as Non-CoT(non reasoning) and CoT(reasoning)}
    \label{fig:result_minif2f}
    \vspace{-1.5em}

\end{figure*}

\textbf{MiniF2F}~\citep{zheng2021minif2f} is a standard benchmark comprising 488 problems from mathematics competitions. Following common practice, we report results on its 244-problem test set, MiniF2F-Test. The results are presented in Figure~\ref{fig:result_minif2f}. EvolProver achieves a pass@32 rate of 69.80\% on MiniF2F-Test, establishing a new SOTA performance among non-reasoning models of comparable size. Notably, this performance is comparable to, and in some cases exceeds, that of reasoning models, despite using significantly fewer tokens (a nearly 10-fold reduction in token consumption).


\textbf{Ineq-Comp}~\citep{zhao2025ineq} is a benchmark designed to evaluate the robustness of formal provers against minor problem perturbations. It contains 75 seed problems from Olympiad-level inequalities and 150 corresponding transformed variants.  
Each seed problem is systematically altered through simple operations(e.g., algebraic rewrites, variable duplication) to create two transformed variants. While humans can easily solve these transformed problems, formal provers often struggle with them even if they can solve the original. A model's robustness is measured by the ratio of its performance on transformed problems to its performance on the seed problems, for which a higher ratio indicates greater robustness.

Our results are presented in \Cref{table:ineq_comp_result}. EvolProver again sets a new SOTA for non-reasoning models on all three metrics (seed, transformed, and ratio), outperforming the next-best non-reasoning model by a significant margin. Its performance is also comparable to that of top reasoning models. Notably, our data augmentation pipeline leads to a substantial boost in robustness: EvolProver's robustness ratio is 30.61 percentage points higher than that of EvolProver-Base, demonstrating the effectiveness of our approach.

\begin{table}
    \centering
    \caption{Comparison with SOTA 7B-size models on the Ineq-Comp Benchmark; Pass means pass@32 rate for reasoning models and non-reasoning models, and means 1 $\times$ 3200 pass rate for tree-search models.\textcolor{red!85!black}{$\uparrow$} means increase in absolute performance over the ablation model EvolProver-Base.}
    \label{table:ineq_comp_result}
    \vspace{0.5\baselineskip}
    \resizebox{0.95\textwidth}{!}{%
        \begin{tabular}{lccc}
        \addlinespace
        \toprule
        \textbf{Models} & \textbf{Pass on Seed} & \textbf{Pass on Transformed} & \textbf{Pass Ratio} \\
        \midrule
        \multicolumn{4}{l}{\textit{Reasoning Models}} \\ 

        ~\makecell[l]{DeepSeek-Prover-V2 (COT)}   & 66.23\% & 44.53\% & 67.23\% \\
        ~Kimina-Prover-Preview      & 50.06\% & 27.58\% & 55.09\% \\
        \midrule

        \multicolumn{4}{l}{\textit{Tree-Search Models}} \\ 

        ~\makecell[l]{DeepSeek-Prover-V1.5 (RL + RMaxTS)}   & 42.66\% & 14.83 \% & 34.76 \% \\

        ~InternLM2.5-StepProver   & 25.59\% & 3.44 \% & 16.6 \% \\

        \midrule
        \multicolumn{4}{l}{\textit{Non-Reasoning Models}} \\ 
        ~\makecell[l]{DeepSeek-Prover-V1.5 (RL)}    & 34.40\% & 6.68\% & 19.42\% \\
        ~Goedel-Prover-SFT          &43.46\% & 14.54\% & 33.47\% \\
        ~STP                        &49.96\% & 18.04\% & 36.12\% \\

        ~\textbf{EvolProver-Base (Ours)}                        & \textbf{43.26\%} & \textbf{14.89\%} & \textbf{34.43\%}\\

        ~\textbf{EvolProver (Ours)}    & \textbf{52.20\%}\textcolor{red!85!black}{\scriptsize{($\uparrow$ 8.94\%)}} & \textbf{34.02\%}\textcolor{red!85!black}{\scriptsize{($\uparrow$ 19.63\%)}} & \textbf{65.17\%}\textcolor{red!85!black}{\scriptsize{($\uparrow$ 30.61\%)}}  \\

        \bottomrule
        \end{tabular}
    }
    \vspace{-1em}
\end{table}

\section{ANALYSIS}
\label{analysis}



\paragraph{Evolution Strategy.} 
EvolDifficulty and EvolDomain employ a general LLM to directly evolve formalized mathematical theorems. This approach addresses the inherent complexity of mathematical formalization, a task traditionally reliant on specialized models trained to convert natural language problems into formal expressions. However, the direct application of general-purpose LLMs for this purpose remains relatively unexplored, leaving their comparative advantages and limitations as an open question.  




To validate our strategy of directly evolving formal statements, we compare it against a common alternative: evolving Natural Language (NL) problems first and then formalizing them. We designed a controlled experiment with four branches:


\begin{itemize}
    \item \textbf{EvolDomain \& EvolDifficulty (Ours)}: Directly evolves new formal statements from existing ones.
    \item \textbf{Formalization-Formalizer}: Evolves NL problems, then formalizes them using a specialized model (Kimina-Formalizer-7B).
    \item \textbf{Formalization-LLM-zero-shot}: Evolves NL problems, then formalizes them using a general-purpose LLM (Gemini-2.5-Pro) in a zero-shot setting.
    \item \textbf{Formalization-LLM-few-shot}: The same as above, but in a few-shot setting.
\end{itemize}

\begin{wrapfigure}{r}{0.5\textwidth} 
    \centering
    \vspace{-1.5em}
    \includegraphics[width=0.8\linewidth]{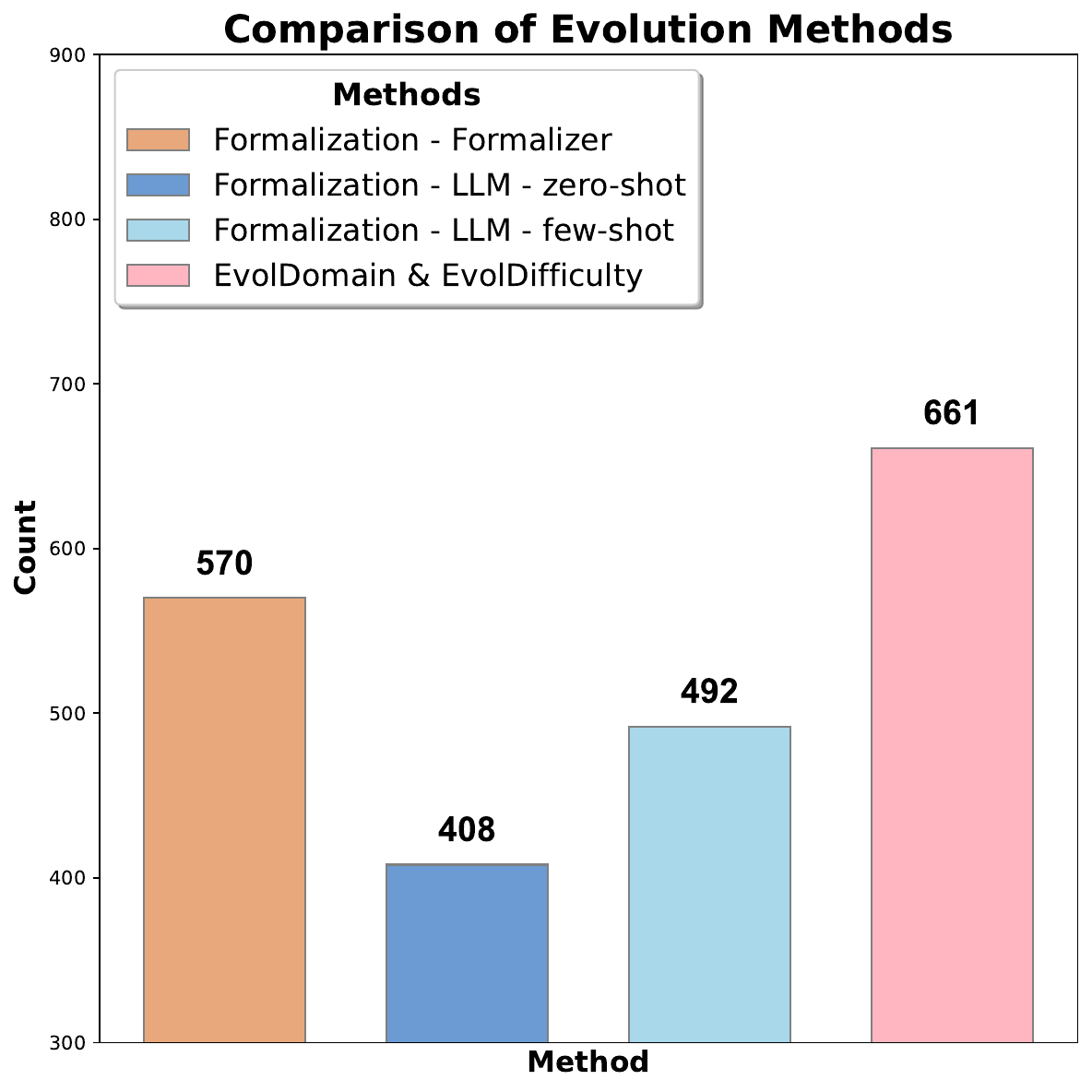}
    \caption{Comparison of the number of candidates passing  verification for four evolution methods. Our EvolDomain \& EvolDifficulty performs best. }
    \label{fig:evol_verified_results}

        \vspace{-1.5em}

\end{wrapfigure}
Starting with 400 seed problems, we generated an equal number of candidates using each method and passed them through our stringent verification pipeline. The number of successfully verified statements for each method is shown in \Cref{fig:evol_verified_results}. Our direct evolution approach significantly outperforms all NL-based methods, confirming its superiority. The final candidate count can exceed 400 as each seed may yield multiple valid variants.

\begin{figure*}[b]
    \centering
    \vspace{-1pt}
    \includegraphics[width=\textwidth]{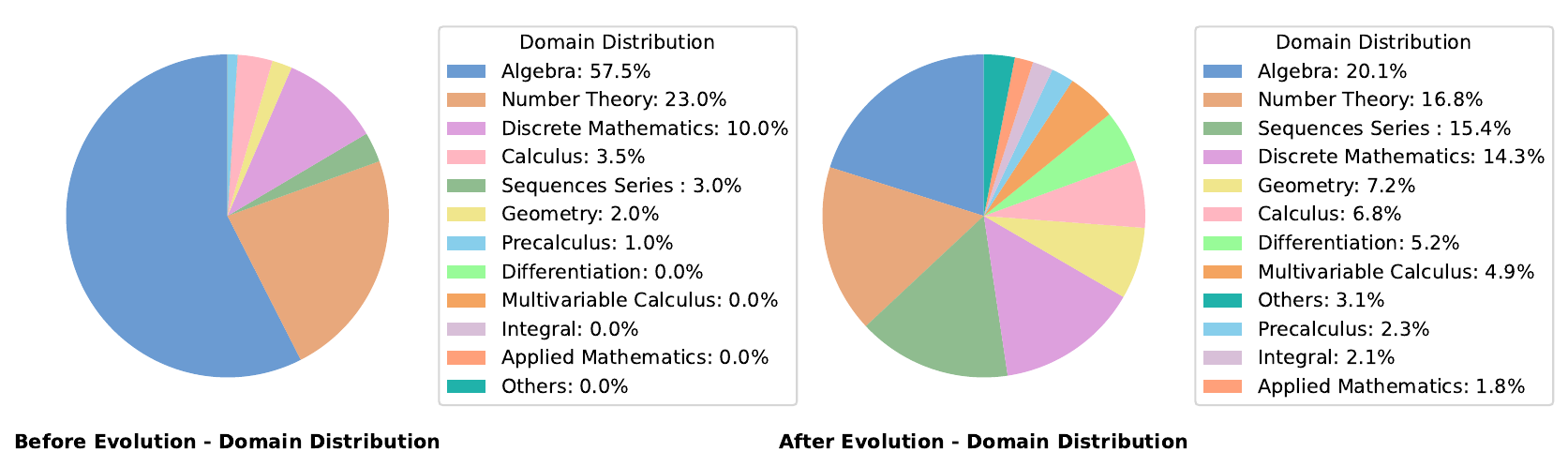}
    \vspace{-1em}
    \caption{Comparison of Mathematical Domain Distribution Before and After EvolDomain.}
    \label{fig:compair_domain}
\end{figure*}

\paragraph{Domain Diversity.} 
Here we analyze how our framework improves domain diversity and how this enhancement translates to performance gains.
\Cref{fig:compair_domain} illustrates the effect of EvolDomain on a sample of 200 seed problems. The initial distribution is heavily skewed, with domains like Algebra dominating while others like Calculus are absent. After applying EvolDomain, the dataset becomes significantly more balanced: the share of over-represented domains is reduced, and previously missing categories are introduced. The domains for both sets were classified by DeepSeek-V3 and human-verified.



This improved diversity directly leads to better model performance across various domains, as detailed in \Cref{table:formalmath_domain}. Comparing EvolProver against the EvolProver-Base baseline, our full model achieves gains across most categories. Critically, it makes a breakthrough in Calculus, solving 3 problems where the baseline solved 0. These results confirm that our strategy not only enriches domain diversity but also enhances the model's overall mathematical capabilities.

%

\begin{table}[t]
\centering
\caption{Number of proved problems on FormalMATH-Lite benchmark in different domains under 32 generation trials. EvolProver improves upon EvolProver-Base across most domains.}
\label{table:formalmath_domain}
\resizebox{0.8\textwidth}{!}{%
    \renewcommand{\arraystretch}{1.0}
    \begin{tabular}{@{\hspace{1.8em}}l@{\hspace{1.8em}}c@{\hspace{1.8em}}c@{\hspace{1.8em}}c@{\hspace{1.8em}}}
    \addlinespace
    \toprule
     \textbf{Domain} & \textbf{EvolProver-Base} & \textbf{EvolProver}  & \textbf{Total}\\
    \midrule
         Algebra & 121 & 141 \textcolor{red!85!black}{(+20)} & 235 \\
    \midrule
         Applied Mathematics & 28 & 33 \textcolor{red!85!black}{(+5)} & 46 \\
    \midrule
         Number Theory & 16 & 23 \textcolor{red!85!black}{(+7)} & 45 \\
    \midrule
         Precalculus & 14 & 15 \textcolor{red!85!black}{(+1)} & 23 \\
    \midrule
         Geometry & 7 & 8 \textcolor{red!85!black}{(+1)} & 17 \\
    \midrule
         Discrete Mathematics & 2 & 5 \textcolor{red!85!black}{(+3)} & 25 \\
    \midrule
         Calculus & 0 & 3 \textcolor{red!85!black}{(+3)} & 6 \\
    \midrule
         Multivariable Calculus & 2 & 2 \textcolor{gray}{(--)} & 6 \\
    \midrule
         Others & 0 & 0 \textcolor{gray}{(--)} & 23 \\
    \bottomrule
  \end{tabular}%
}
\end{table}


\paragraph{Ablation Experiments.}

To further validate the effectiveness of our proposed methods, we conduct a series of comprehensive ablation studies. The results are presented in \Cref{table:ablation_experiments}. These experiments isolate the impact of each component and demonstrate that they provide consistent benefits across multiple benchmarks and at various training stages. Experimental details are provided in Appendix~\ref{subsec:Ablation_Experiments}.

\begin{table}[tb]
\centering
\caption{Ablation experiment results on the FormalMATH-Lite benchmark, the MiniF2F-Test benchmark, and the Ineq-Comp benchmark. All results are pass@32 rate.  Superscripts denote the training data used: superscript\textsuperscript{0} for Public dataset only; superscript\textsuperscript{0+1} for Public dataset + EvolDomain \& EvolDifficulty augmentation; superscript\textsuperscript{0+1+2} for Full augmentation including Public dataset, EvolDomain \& EvolDifficulty, and EvolAST. EvolProver-Ablation-SFT and Evoler-SFT are trained through a sole SFT stage. EvolProver-Base, EvolProver-Ablation-RL and EvolProver are trained through an SFT stage and an RL stage. }
\label{table:ablation_experiments}
\vspace{0.5\baselineskip}

\resizebox{\textwidth}{!}{\begin{tabular}{lcccccc} 
\addlinespace
\toprule
\textbf{Models} & \textbf{FormalMATH}  & \textbf{MiniF2F} & \makecell[c]{\textbf{Ineq-Comp} \\ \textbf{(Seed)}} & \makecell[c]{\textbf{Ineq-Comp} \\ \textbf{(Transformed)}} & \makecell[c]{\textbf{Ineq-Comp} \\ \textbf{(Ratio)}} \\
\midrule
\addlinespace


EvolProver-Base\textsuperscript{0}  & 44.71\% &  52.05\%  & 43.26\% & 14.89\% & 34.43\% \\
\addlinespace


EvolProver-Ablation-SFT\textsuperscript{ 0+1}  & 50.35\% & 65.16\% & 49.79\% & 29.19\% & 58.62\% \\ 
\addlinespace


EvolProver-SFT\textsuperscript{ 0+1+2}  & 51.53\% & 66.39\% & 49.82\% & 30.35\%  & 60.19\% \\
\addlinespace

EvolProver-Ablation-RL\textsuperscript{ 0+1}  & 51.98\% & 68.22\% & 50.36\% & 33.05\% & 65.62\%\\
\addlinespace

\textbf{EvolProver}\textsuperscript{ 0+1+2} & \textbf{53.96\%} & \textbf{69.80\%} & \textbf{52.20\%} & \textbf{34.02\%}  & \textbf{65.17\%}\\

\bottomrule
\end{tabular}}
\end{table}

\section{Conclusion and Future Work}
\label{conclusion_and_future_work}

In this paper, we introduced a highly-extensible data augmentation pipeline with three methods: EvolDomain, EvolDifficulty, and EvolAST, designed to improve model generalizability from semantic and syntactic perspectives. Our resulting model, EvolProver, achieves new SOTA results on several key benchmarks, notably surpassing all comparable models on FormalMATH-Lite. For future work, we plan to enhance EvolProver's reasoning capabilities by incorporating synthetically generated Chain-of-Thought data into its training.

\section*{Reproducibility Statement}
We are strongly committed to the reproducibility of our work. Our EvolAST method is designed to be highly extensible, and we encourage the community to contribute by expanding its set of applicable axioms and theorems. To facilitate this, we will release our code and models publicly upon receiving institutional approval.

\section*{Acknowledgment}
This work was supported by Ant Group Research Intern Program.

\bibliography{iclr2026_conference}
\bibliographystyle{iclr2026_conference}

\clearpage

\appendix
\section{Appendix}
\subsection{Details of Training EvolProver}
\label{subsec:training_evolprover}
\paragraph{Data Curation.}
Our data curation process follows a multi-stage funnel. We begin with a seed pool of approximately 3.3 million verified formal statements aggregated from four sources: DeepSeek-Prover-V1, STP-lean, MiniF2F-Valid, and FormalMATH-All (excluding the FormalMATH-Lite subset).

From this pool, we sample 70k statements for evolution. These are processed by our EvolDomain and EvolDifficulty methods using Gemini-2.5-Pro and DeepSeek-R1, which then undergo a verification process to yield 57.4k high-quality (statement, description) pairs. This verification first involves a syntax check using the Lean 4 compiler; if the compiler finds a syntax error, we use DeepSeek-V3~\citep{deepseek_v3} to repair it, after which DeepSeek-V3 performs a final semantic check. Next, we apply EvolAST to this set for syntactic diversification, expanding it to approximately 96.7k entries (as a single statement can generate multiple AST variants). Finally, we generate proofs for each statement. Using DeepSeek-Prover-V2-671B and Goedel-Prover-V2-8B as expert models, we generate 50 proof candidates per statement and retain only those that pass Lean 4 compiler verification. After removing duplicates, this process results in a final training dataset of 39.2k unique (statement, proof) pairs. To prevent data leakage,  We ensure that the initial states of all theorem statements in our data are different from those in the tested benchmarks. 

\paragraph{Supervised Fine Tuning.} We fine-tune the DeepSeek-Prover-V1.5-Base model using full-parameter supervised fine-tuning (SFT). Our training data is a mixture of our self-evolved instruction dataset and publicly available datasets. The model is trained for one epoch with the AdamW optimizer. We set the initial learning rate to 
$1.0 \times 10^{-5}$ and decay it  using a cosine scheduler with a 5\% warmup ratio. All sequences are truncated to a maximum length of 4096 tokens, and we use a global batch size of 32.

\textbf{Reinforcement Learning.} Following the Supervised Fine-Tuning (SFT) stage, we further enhance the model's performance by applying Reinforcement Learning (RL) to the SFT checkpoint. For RL training, we utilize our augmented open-source dataset of formal problems. We employ a standard binary reward: for each problem, the model receives a reward of $1$ if the generated Lean proof is correct, and $0$ otherwise. This RL fine-tuning process produces the final EvolProver.

\paragraph{RL Training Details} To improve training efficacy, we curate the RL training dataset by filtering problems based on the pass@1 success rate of the SFT checkpoint. We include only problems where $0 < \text{pass}@1 < 1/2$. This selection strategy ensures that the training set is challenging yet solvable for our model. The filtered dataset contains 2,718 problems. We initialize both the actor and critic models with the weights from the SFT checkpoint and train them using Proximal Policy Optimization (PPO). The training runs for 10 epochs with a batch size of 256, a constant actor learning rate of $1.0 \times 10^{-6}$, a constant critic learning rate of $1.0 \times 10^{-5}$, a clip ratio of $0.2$, and a KL divergence loss coefficient of $0.001$.

\subsection{Ablation Experiments}
\label{subsec:Ablation_Experiments}



\subsubsection{Ablation Model Training}
To precisely evaluate the contribution of each component, we trained a series of ablation models under controlled conditions. All training hyperparameters were kept identical across corresponding stages. The models are:

\begin{itemize}
    \item \textbf{EvolProver-Base}: Our baseline, trained on the original, unaugmented public dataset through both SFT and RL stages.
    
    \item \textbf{EvolProver-Ablation-SFT}: Trained on data augmented only by EvolDomain and EvolDifficulty, and only undergoes the SFT stage.
    
    \item \textbf{EvolProver-Ablation-RL}: Same data as above (EvolDomain and EvolDifficulty only), but undergoes the full SFT and RL training process. This model directly isolates the impact of EvolAST when compared to the final EvolProver.
    
    \item \textbf{EvolProver-SFT}: The checkpoint of our final model after being trained on the fully augmented dataset (including EvolAST) for the SFT stage only.
\end{itemize}

We did not create an ``EvolAST-only" model, as EvolAST operates on the output of EvolDomain and EvolDifficulty, making such an experiment logically infeasible.



%

\subsubsection{Ablation Experiment Results}
The results of our ablation experiments, presented again in \Cref{table:ablation_experiments_in_appendix} for convenience, lead to two key conclusions.
First, data augmentation provides a substantial boost, with even the partially augmented models (Ablation-SFT/RL) drastically outperforming the EvolProver-Base model across all benchmarks, often by more than 10\%.
Second, the EvolAST method consistently yields further improvements across all benchmarks. In the SFT stage, EvolProver-SFT (with EvolAST) surpasses EvolProver-Ablation-SFT (without EvolAST). Similarly, in the RL stage, the final EvolProver outperforms EvolProver-Ablation-RL. This demonstrates the value of the EvolAST method in both training phases.

\begin{table}[ht]
\centering
\caption{Ablation experiment results on the FormalMATH-Lite benchmark, the MiniF2F-Test benchmark, and the Ineq-Comp benchmark. All results are pass@32 rate.  Superscripts denote the training data used: superscript\textsuperscript{0} for Public dataset only; superscript\textsuperscript{0+1} for Public dataset + EvolDomain \& EvolDifficulty augmentation; superscript\textsuperscript{0+1+2} for Full augmentation including Public dataset, EvolDomain \& EvolDifficulty, and EvolAST.}
\label{table:ablation_experiments_in_appendix}
\vspace{0.5\baselineskip}

\resizebox{\textwidth}{!}{\begin{tabular}{lcccccc} 
\addlinespace
\toprule
\textbf{Models} & \textbf{FormalMATH}  & \textbf{MiniF2F} & \makecell[c]{\textbf{Ineq-Comp} \\ \textbf{(Seed)}} & \makecell[c]{\textbf{Ineq-Comp} \\ \textbf{(Transformed)}} & \makecell[c]{\textbf{Ineq-Comp} \\ \textbf{(Ratio)}} \\
\midrule
\addlinespace


EvolProver-Base\textsuperscript{0}  & 44.71\% &  52.05\%  & 43.26\% & 14.89\% & 34.43\% \\
\addlinespace


EvolProver-Ablation-SFT\textsuperscript{ 0+1}  & 50.35\% & 65.16\% & 49.79\% & 29.19\% & 58.62\% \\ 
\addlinespace


EvolProver-SFT\textsuperscript{ 0+1+2}  & 51.53\% & 66.39\% & 49.82\% & 30.35\%  & 60.19\% \\
\addlinespace

EvolProver-Ablation-RL\textsuperscript{ 0+1}  & 51.98\% & 68.22\% & 50.36\% & 33.05\% & 65.62\%\\
\addlinespace

\textbf{EvolProver}\textsuperscript{ 0+1+2} & \textbf{53.96\%} & \textbf{69.80\%} & \textbf{52.20\%} & \textbf{34.02\%}  & \textbf{65.17\%}\tablefootnote{While the performance ratio for EvolProver is slightly lower than that of EvolProver-Ablation-RL, this metric can be misleading when viewed in isolation. This is because EvolProver demonstrates absolute gains in both the numerator and the denominator of the fraction. The marginal decrease in the final ratio is therefore an artifact of the denominator's more substantial growth, rather than an indication of inferior performance.}\\

\bottomrule
\end{tabular}}

\end{table}

\clearpage
\clearpage
\newenvironment{evoldomain_prompt}{%
  \begin{tcolorbox}[colback=beaublue!8!white,breakable,colframe=beaublue!10!black,title=Prompt Template for EvolDimain]
}{%
  \end{tcolorbox}
}

\newenvironment{original_statement}{%
  \begin{tcolorbox}[colback=brown!8!white,colframe=brown!10!black,breakable,title=Original Lean 4 Statement (Number Theory)]
}{%
  \end{tcolorbox}
}

\newenvironment{evolved_statement_geometry}{%
  \begin{tcolorbox}[colback=brown!8!white,colframe=brown!10!black,title=Evolved Statement and its corresponding Natural Language Question (Geometry) ]
}{%
  \end{tcolorbox}
}

\newenvironment{evolved_statement_integral}{%
  \begin{tcolorbox}[colback=brown!8!white,colframe=brown!10!black,title=Evolved Statement and its corresponding Natural Language Question (Integral) ]
}{%
  \end{tcolorbox}
}

\subsection{DETAILS FOR EvolDomain}
\label{appendix:details_for_evoldomain}

\subsubsection{Prompt Template}
Formal problems can precisely extract the universal logical skeleton of a mathematical problem. Our strategy leverages this by transferring that structure to new domains to systematically create rigorous new problems. Our preset domains cover a range of topics from high school competition problems to undergraduate-level subjects. The prompt template format of EvolDomain is as follow:

\begin{evoldomain_prompt}
Your task is to start with a given Lean 4 formalized problem and follow the strategy below to formulate a new problem in a different mathematical domain.
\\\\
 \#\#\# Transformation Strategy \\ \\
 Step 1. Deconstruction \& Abstraction  \\
 Identify the original statement's abstract logical skeleton by isolating its core components. This involves recognizing the underlying mathematical objects, the essential operations being performed, and the fundamental relationship being asserted. \\\\
 Step 2. Analogy \& Transfer  \\\\
 Find a parallel structure in a new mathematical domain by identifying an analogous sequence of objects in the list below.

[``Algebra", ``Number Theory", ``Integral", ``Precalculus", ``Differentiation", ``Multivariable Calculus", ``Sequences Series", ``Applied Mathematics", ``Discrete Mathematics", ``Geometry", ``Calculus", ``Other"]

 Then, translate the original operations and relations into concepts that are natural within this new context. \\\\
 Step 3. Instantiation \& Packaging  \\\\
 Formulate a new, concrete problem by applying the transferred structure and operations to the analogous objects. Package this new proposition into a clear natural language description and a rigorous formal statement. 
\\\\
 Evolved variants should follow the following requirements: \\  
 1. Each variant should be mathematically rigorous and meaningful  \\
 2. Each variant should be syntax correct and a valid Lean 4 statement  \\
 3. Each variant should differ from the original mathematical domain\\
 4. Each variant should follow the same output format as ``\#\#\# Example Variant Format"  
\\\\
 \#\#\# Example Variant Format: \\ 
 \verb|```|NL Description  \\
 Prove that ...  \\
 \verb|```|  \\
 \verb|```|Formal Statement  \\
 ... := by sorry  \\
 \verb|```|  \\
\verb|```| Domain\\
Target Domain in the list\\
\verb|```|\\

 Please provide 3-5 variants following the strategies and requirements above, based on original statement. \\

   \#\#\# Original Formal Statement \\ 
 \verb|```|lean4  \\
 \{Original Formal Statement\}\\
 \verb|```|    \\\\
 \#\#\# Evolution Results (3-5 variants in other mathematical domains)
\end{evoldomain_prompt}

\subsubsection{Case Study}

We select an example of EvolDomain from the evolved dataset. The original Lean 4 statement is as follows:
\begin{original_statement}
\begin{minted}[escapeinside=||, texcl=true]{lean}
theorem lean_workbook_12011 |$(m, n : \mathbb{Z})$| (h_1 : 2*m + n = 0) : m|\textasciicircum|3 |$\neq$| n|\textasciicircum|3 - 15 := by sorry
\end{minted}
\end{original_statement}


The evolved Lean 4 statement and its corresponding natural language description are as follows:

\begin{evolved_statement_geometry}
    
Evolved Formal Statement:
\vspace{-10pt} 
    \begin{minted}[escapeinside=||, texcl=true]{lean}
theorem affine_points_analogy P Q R : EuclideanSpace |$\mathbb{R}$| (Fin 2)) (h : (2 |$\cdot$| (Q - P)) + (R - Q) = 0) : dist P Q |\textasciicircum| 2 |$\neq$| dist P R |\textasciicircum| 2 - 15 := by sorry
\end{minted}
\vspace{5pt} 
Natural Language Question:
\vspace{-10pt} 
\begin{minted}[escapeinside=||, texcl=true]{lean}
Consider three points `P`, `Q`, and `R` |in| a plane. If the vector |from| `Q` to `R` added to twice the vector |from| `P` to `Q` results |in| the zero vector, prove that the squared distance |from| `P` to `Q` is not equal to the squared distance |from| `P` to `R` minus 15.
\end{minted}
\end{evolved_statement_geometry}

\begin{evolved_statement_integral}
    Evolved Formal Statement:
\vspace{-10pt} 
\begin{minted}[escapeinside=||, texcl=true]{lean}
theorem integral_variant_1
  (f g : |$\mathbb{R}  \rightarrow \mathbb{R}$|)
  (hf : ContinuousOn f (Set.Icc 0 1))
  (hg : ContinuousOn g (Set.Icc 0 1))
  (h_f_nonneg : |$\forall$| x |$\in$| Set.Icc 0 1, f x |$\ge$| 0)
  (h_f_not_zero : |$\exists$| x |$\in$| Set.Icc 0 1, f x |$\neq$| 0)
  (h_1 : |$\forall$| x |$\in$| Set.Icc 0 1, 2 * f x + g x = 0) :
  |$\int$| x in (0)..1, (f x)|\textasciicircum|3 |$\neq$| |$\int$| x in (0)..1, (g x)|\textasciicircum|3 - 15 := by sorry
\end{minted}

\vspace{5pt} 
Natural Language Question:
\vspace{-10pt} 
\begin{minted}[escapeinside=||, texcl=true]{lean}
Prove that for any two real-valued continuous functions, `f` and `g`, defined on the interval `[0, 1]`, |if| `f` is non-negative and not identically zero, and |if| `2 |*| f(x) + g(x) = 0` for all `x` |in| `[0, 1]`, |then| the integral of `f(x)|\textasciicircum|3` over `[0, 1]` is not equal to the integral of `g(x)|\textasciicircum|3` over `[0, 1]` minus 15.
\end{minted}
\end{evolved_statement_integral}

\newenvironment{evoldifficulty_prompt_increase}{%
  \begin{tcolorbox}[colback=beaublue!8!white,breakable,colframe=beaublue!10!black,title=Prompt Template for EvolDifficulty (Increase Difficulty)]
}{%
  \end{tcolorbox}
}

\newenvironment{evoldifficulty_prompt_decrease}{%
  \begin{tcolorbox}[colback=beaublue!8!white,breakable,colframe=beaublue!10!black,title=Prompt Template for EvolDifficulty (Decrease Difficulty)]
}{%
  \end{tcolorbox}
}

\newenvironment{evoldifficulty_strategy_increase}{%
  \begin{tcolorbox}[colback=beaublue!8!white,breakable,colframe=beaublue!10!black,title=Strategies and Specific Methods (Increase Difficulty)]
}{%
  \end{tcolorbox}
}

\newenvironment{evoldifficulty_strategy_decrease}{%
  \begin{tcolorbox}[colback=beaublue!8!white,breakable,colframe=beaublue!10!black,title=Strategies and Specific Methods (Decrease Difficulty)]
}{%
  \end{tcolorbox}
}

\newenvironment{original_statement_difficulty_1}{%
  \begin{tcolorbox}[colback=brown!8!white,colframe=brown!10!black,breakable,title=Original Lean 4 Statement for Upward Evolution]
}{%
  \end{tcolorbox}
}

\newenvironment{Evolved_statement_difficulty_1}{%
  \begin{tcolorbox}[colback=brown!8!white,colframe=brown!10!black,breakable,title= Evolved Statement and its corresponding Natural Language Question (Increase Difficulty)]
}{%
  \end{tcolorbox}
}

\newenvironment{original_statement_difficulty_2}{%
  \begin{tcolorbox}[colback=brown!8!white,colframe=brown!10!black,breakable,title=Original Lean 4 Statement for Downward Evolution]
}{%
  \end{tcolorbox}
}

\newenvironment{Evolved_statement_difficulty_2}{%
  \begin{tcolorbox}[colback=brown!8!white,colframe=brown!10!black,breakable,title= Evolved Statement and its corresponding Natural Language Question (Decrease Difficulty)]
}{%
  \end{tcolorbox}
}

\subsection{DETAILS FOR EvolDifficulty}
\label{appendix:details_for_evoldifficulty}
To generate problems of varying difficulty, we define a set of evolution strategies through deliberation and screening by a team of mathematics experts. These strategies fall into two categories: upward (to increase difficulty) and downward (to decrease difficulty), each with specific methods for implementation.
\subsubsection{UpwardEvolution}
The prompt tempalte for upward evolution is as follows:
\begin{evoldifficulty_prompt_increase}
    
Your task is to evolve a given formal statement into several, more complex formal statements, according to the provided strategies and requirements. For each new formal statement, you must provide its corresponding natural language meaning.
\\
\#\#\# Difficulty Enhancement Strategy\\

Your objective is to \{strategy\} for the original statement. 
\\\\
First, understand the core concept and structure of the original formal statement. Identify its key logical components, such as variables, propositions, logical operators, quantifiers, conditions, and the overall scope. Then, you can select from a range of strategies, including but not limited to the following, to enhance difficulty:
\\
\{Specific Methods\}

...
\\
\#\#\# Evolution Requirements

Evolved variants should follow the following requirements:\\
1. Each variant must represent a genuine enhancement of its proof's logic and difficulty, not just an increase in superficial complexity.\\
2. Each variant should be mathematically rigorous and meaningful\\
3. Each variant should be syntax correct and a valid Lean 4 statement\\
4. Each variant should be different from the original statement and other variants\\
5. Each variant should follow the same output format as ``\#\#\# Example Variant Format".\\

\#\#\# Example Variant Format:\\
\verb|```|NL Description\\
Prove that ...\\
\verb|```|\\
\verb|```|Formal Statement\\
... := by sorry\\
\verb|```|\\
\\
Please provide 3-5 variants following the strategies and requirements.\\

\#\#\# Original Formal Statement\\
\verb|```|lean4\\
\{Original Formal Statement\}\\
\verb|```|\\

\#\#\# Evolution Results (3-5 variants with increasing difficulty)
\end{evoldifficulty_prompt_increase}

The strategies and specific methods are as follows:
\begin{evoldifficulty_strategy_increase}
    1. Complicate the Logical Structure\\
(1) Construct a new problem that increases the nesting depth and layers of the original problem's propositional logic.\\
(2) Construct a new problem by introducing a logical system with complex dependencies between its components.\\
(3) Construct a new problem whose internal structure is obscured by multiple layers of non-obvious equivalent transformations.\\

2. Increase the Mathematical Depth\\
(1) Construct a new problem that relies on a deeper theoretical framework.\\
(2) Construct a new problem that requires a longer, but logically similar, chain of reasoning to solve.\\
(3) Construct a new problem that positions the original problem as a critical sub-problem or lemma within its proof.\\

3. Elevate Abstraction and Generalization\\
(1) Construct a new problem by elevating and generalizing a specific instance or special case from the original problem into a universal proposition that must be proven.\\
(2) Construct a new problem that adds stricter conditions, requiring reasoning and verification under them.\\
(3) Construct a new problem whose proof requires the fusion of concepts or tools from different knowledge domains.\\

4. Intensify Constraints and Precision\\
(1) Construct a new problem that increases complexity by establishing critical boundaries or singularities within the problem's domain.\\
(2) Construct a new problem that adds specific, strong constraints, requiring the discovery of an optimal solution or an extremal state.\\
(3) Construct a new problem with heightened rigor requirements, making it necessary to provide a strict argument for the existence, uniqueness, or enumeration of the solution(s).\\

5. Add Parametric and Analytical Complexity\\
(1) Construct a new problem that broadens the hypothesis space and increases analytical complexity by introducing or adjusting explicit parameters.\\
(2) Construct a new problem whose internal structure spans both discrete and continuous forms, requiring a transformation between them (e.g., the limit relationship between a sum and an integral) to be solved.
\end{evoldifficulty_strategy_increase}

\subsubsection{DownwardEvolution}
The prompt template for downward evolution is as follows:
\begin{evoldifficulty_prompt_decrease}
Your task is to evolve a given formal statement into several, simpler formal statements, according to the provided strategies and requirements. For each new formal statement, you must provide its corresponding natural language meaning.
\\
\#\#\# Difficulty Reduction Strategy\\

Your objective is to \{strategy\} for the original statement.
\\\\
First, understand the core concept and structure of the original formal statement. Identify its key logical components, such as variables, propositions, logical operators, quantifiers, conditions, and the overall scope. Then, you can select from a range of strategies, including but not limited to the following, to reduce difficulty:
\\
\{Specific Methods\}

...
\\
\#\#\# Evolution Requirements

Evolved variants should follow the following requirements:\\
1. Each variant must represent a genuine simplification of its proof's logic and structure, not just a cosmetic rephrasing.\\
2. Each variant should be mathematically rigorous and meanigful\\
3. Each variant should be syntax correct and a valid Lean 4 statement\\
4. Each variant should be different from the original statement and other variants\\
5. Each variant should follow the same output format as ``\#\#\# Example Variant Format".\\

\#\#\# Example Variant Format:\\
\verb|```|NL Description\\
Prove that ...\\
\verb|```|\\
\verb|```|Formal Statement\\
... := by sorry\\
\verb|```|\\
\\
Please provide 3-5 variants following the strategies and requirements.\\

\#\#\# Original Formal Statement\\
\verb|```|lean4\\
\{Original Formal Statement\}\\
\verb|```|\\

\#\#\# Evolution Results (3-5 variants with decreasing difficulty)
\end{evoldifficulty_prompt_decrease}

The strategies and specific methods are as follows:
\begin{evoldifficulty_strategy_decrease}
    1. Simplify the Logical Structure\\
(1) Construct a new problem that decreases the nesting depth and layers of the proposition's logic.\\
(2) Construct a new problem containing a logical system with weaker or no dependencies between its components.\\
(3) Construct a new problem whose internal structure is transparent, solvable through direct logical relations rather than non-obvious transformations.\\

2. Reduce the Mathematical Depth\\
(1) Construct a new problem that relies on a more elementary theoretical framework.\\
(2) Construct a new problem that only requires completing the initial steps or the final conclusion of the original problem's longer reasoning chain.\\
(3) Construct a new problem by isolating a key lemma or an intermediate step from the original problem's proof and setting it as the sole objective.\\

3. Reduce Abstraction and Specialize\\
(1) Construct a new problem by taking a general or abstract proposition and creating a specific, concrete instance of it to be solved or verified.\\
(2) Construct a new problem that replaces abstract symbols and variables with concrete numerical values or tangible examples to lower the barrier to understanding.\\
(3) Construct a new problem by reformulating it so that it can be solved using concepts and tools from a single, self-contained knowledge domain, avoiding interdisciplinary fusion.\\

4. Loosen Constraints and Precision\\
(1) Construct a new problem by restricting its domain to regular cases, excluding critical boundaries or singularities.\\
(2) Construct a new problem that requires finding any feasible solution rather than an optimal or extremal one.\\
(3) Construct a new problem that asks for a single concrete example of a solution, rather than a rigorous proof of its existence, uniqueness, or enumeration.\\

5. Reduce Parametric and Analytical Complexity\\
(1) Construct a new problem that reduces the dimension of analysis by reducing the number of variables required to address the problem or by simplifying a complex functional relationship between parameters to a linear one.\\
\end{evoldifficulty_strategy_decrease}

\subsubsection{Case Study}

We select two examples of EvolDifficulty from evolved dataset. The original Lean 4 statement for upward evolution is as follows:

\begin{original_statement_difficulty_1}
\begin{minted}[escapeinside=##, texcl=true]{lean}
theorem lean_workbook_37427 #$(n:\mathbb{N})$# : 133 | 11#\textasciicircum#(n+2) + 12#\textasciicircum#(2*n+1) := by sorry
\end{minted}
\end{original_statement_difficulty_1}

The evolved Lean 4 statement and its corresponding natural language description are as follows:

\begin{Evolved_statement_difficulty_1}
Evolved Formal Statement:
\vspace{-10pt} 
    \begin{minted}[escapeinside=##, texcl=true]{lean}
theorem variant_1 #$(n : \mathbb{N})$# : 133 | Nat.gcd (11#\textasciicircum#(n+2) + 12#\textasciicircum#(2*n+1)) (11#\textasciicircum#((n+1)+2) + 12#\textasciicircum#(2*(n+1)+1)) := by sorry
\end{minted}
\vspace{5pt} 
Natural Language Question:
\vspace{-10pt} 
\begin{minted}[escapeinside=||, texcl=true]{lean}
Prove that for any natural number n, the greatest common divisor of 11|\textasciicircum|(n+2) + 12|\textasciicircum|(2*n+1) and 11|\textasciicircum|((n+1)+2) + 12|\textasciicircum|(2*(n+1)+1) is divisible |by| 133.
\end{minted}
\end{Evolved_statement_difficulty_1}

\begin{Evolved_statement_difficulty_1}
Evolved Formal Statement:
\vspace{-10pt} 
    \begin{minted}[escapeinside=##, texcl=true]{lean}
def u_seq : #$\mathbb{N} \rightarrow \mathbb{Z}$#
  | 0 => 133
  | 1 => 3059
  | k + 2 => 155 * u_seq (k + 1) - 1584 * u_seq k
theorem variant_2 #$n : \mathbb{N}$# : 133 | u_seq n := by sorry
\end{minted}
\vspace{5pt} 
Natural Language Question:
\vspace{-10pt} 
\begin{minted}[escapeinside=||, texcl=true]{lean}
Let an integer sequence u be defined |by| u_0 = 133, u_1 = 3059, and the recurrence relation u_{n+2} = 155 * u_{n+1} - 1584 * u_n for all n |$\ge$| 0. Prove that every term u_n of the sequence is divisible |by| 133.
\end{minted}
\end{Evolved_statement_difficulty_1}

The original Lean 4 statement for downward evolution is as follows:

\begin{original_statement_difficulty_2}
\begin{minted}[escapeinside=||, texcl=true]{lean}
theorem lean_workbook_21498 (a b : |$\mathbb{R}$|) (ha : 0 < a) (hb : 0 < b) : (a + b) |\textasciicircum| 2 / 2 + (a + b) / 4 |$\ge$| a * Real.sqrt b + b * Real.sqrt a := by sorry
\end{minted}
\end{original_statement_difficulty_2}

\newpage

The evolved Lean 4 statement and its corresponding natural language description are as follows:

\begin{Evolved_statement_difficulty_2}
Evolved Formal Statement:
\vspace{-10pt} 
    \begin{minted}[escapeinside=||, texcl=true]{lean}
theorem variant_1 |$(a, b : \mathbb{R})$| (ha : 0 |$\le$| a) (hb : 0 |$\le$| b) : (Real.sqrt a + Real.sqrt b) |\textasciicircum|2  |$\le$| 2 * (a + b) := by sorry
\end{minted}
\vspace{5pt} 
Natural Language Question:
\vspace{-10pt} 
\begin{minted}[escapeinside=||, texcl=true]{lean}
Prove that for any non-negative real numbers `a` and `b`, the square of the sum of their square roots is less than or equal to twice their sum.
\end{minted}
\end{Evolved_statement_difficulty_2}

\begin{Evolved_statement_difficulty_2}
Evolved Formal Statement:
\vspace{-10pt} 
    \begin{minted}[escapeinside=||, texcl=true]{lean}
theorem variant_2 |$(a, b : \mathbb{R})$| (ha : 0 < a) (hb : 0 < b) : (a + b) / 2 |$\ge$| Real.sqrt (a * b) := by sorry
\end{minted}
\vspace{5pt} 
Natural Language Question:
\vspace{-10pt} 
\begin{minted}[escapeinside=||, texcl=true]{lean}
Prove that Arithmetic Mean-Geometric Mean (AM-GM) inequality for two positive real numbers `a` and `b`, which states that their arithmetic mean is greater than or equal to their geometric mean.
\end{minted}
\end{Evolved_statement_difficulty_2}
\newenvironment{consistency_prompt}{%
  \begin{tcolorbox}[colback=beaublue!8!white,breakable,colframe=beaublue!10!black,title=Prompt Template to Judge Consistency]
}{%
  \end{tcolorbox}
}

\newenvironment{correctness_prompt}{%
  \begin{tcolorbox}[colback=beaublue!8!white,breakable,colframe=beaublue!10!black,title=Prompt Template to Judge Correctness]
}{%
  \end{tcolorbox}
}

\newenvironment{difficulty_prompt}{%
  \begin{tcolorbox}[colback=beaublue!8!white,breakable,colframe=beaublue!10!black,title=Prompt Template for Filtering Out Low-difficulty Problems]
}{%
  \end{tcolorbox}
}

\newenvironment{modify_prompt}{%
  \begin{tcolorbox}[colback=beaublue!8!white,breakable,colframe=beaublue!10!black,title=Prompt Template for Correcting Formal Statement]
}{%
  \end{tcolorbox}
}

\subsection{DETAILS FOR Verification}
\label{appendix:details_for_verification}

The prompt template to judge consistency between natural language problem and formal statement is as follows:

\begin{consistency_prompt}
You will be provided with a Natural Language Description and a Formal Statement. Please judge if they are consistent, and provide specific analysis:\\

Natural Language Description:\\
\verb|```|Problem\\
\{Natural Language Description\}\\
\verb|```|\\
Formal Statement:\\
\verb|```|lean4\\
\{Formal Statement\}\\
\verb|```|\\

For your response, please follow this example format:\\
**Consistency Analysis:**\\
\verb|```|analysis\\
Your detailed analysis\\
\verb|```|\\
**Judge Result:**\\
\verb|```|judge\\
Consistent or Inconsistent\\
\verb|```|\\

Now, please provide your formal answer:
\end{consistency_prompt}

The prompt template to judge mathematical correctness of formal statements and natural language problem is as follows:

\begin{correctness_prompt}
    You will be provided with a Natural Language Description and a Formal Statement. Please judge if the mathematical statement is correct, and provide specific analysis:\\

Natural Language Description:\\
\verb|```|Problem\\
\{original nl\}\\
\verb|```|\\

Formal Statement:\\
\verb|```|lean4\\
\{correct formal statement\}\\
\verb|```|\\

Please analyze the mathematical correctness by considering:\\
1. Whether the problem is provable (can be proven or disproven)\\
2. Whether the problem statement is well-formed and meaningful\\
3. Whether there are any logical contradictions or inconsistencies\\

For your response, please follow this example format:\\
**Mathematical Correctness Analysis:**\\
\verb|```|analysis\\
Your detailed analysis\\
\verb|```|\\
**Judge Result:**\\
\verb|```|judge\\
Correct or Incorrect\\
\verb|```|\\

Now, please provide your formal answer:
\end{correctness_prompt}

The prompt template for filtering out low-difficulty problems is as follows:
\begin{difficulty_prompt}
You will be provided with a Natural Language Description and a Formal Statement. Your task is to classify the difficulty of problem in Lean 4:\\

Natural Language Description:\\
\verb|```|Problem\\
\{Natural Language Description\}\\
\verb|```|\\
Formal Statement:\\
\verb|```|lean4\\
\{Formal Statement\}\\
\verb|```|\\

Please analyze the problem and determine if it is Low-difficulty. Here are the criteria for a Low-difficulty problem:\\
1. Simple calculations\\
2. Simple algebraic manipulations \\ 
3. Solving single variable linear equations (by just a 1-step calculation)\\
4. Inequalities proved by an easy sum-of-squares technique\\

Conversely, the following types of problems should NOT be classified as Low-difficulty:\\
1. Inequality proving with the square root (might be more complex)\\
2. More complex inequalities, limits, and integrals\\
3. Problems dealing with integers (more related to number theory)\\
4. Problems involving higher order roots, complex numbers, matrices, polynomials, group, finite-sum, or functional equations (since these might shed lights on other hard problems)\\

For your response, please follow this example format:

**Difficulty Analysis:**\\
\verb|```|analysis\\
Your detailed analysis\\
\verb|```|\\
**Judge Result:**\\
- Is Low-difficulty: \\
\verb|```|judge\\
Yes or No\\
\verb|```|\\

Now, please provide your formal answer:
\end{difficulty_prompt}

Prompt template for fixing compilation errors in a formal statement is as follows:

\begin{modify_prompt}
 Your task is to fix the code based on the errors and provide a corrected version. Please also provide a detailed analysis of the changes you made. You will be provided with an incorrect Lean4 code snippet and a list of corresponding errors.\\

Incorrect Lean4 Code:\\
\verb|```|lean4\\
{incorrect lean4 code}\\
\verb|```|\\
Error List:\\
\verb|```|errors\\
{errors}\\
\verb|```|\\

Please modify the incorrect Lean 4 code according to the following requirements:\\
1.The corrected statement must be syntactically valid and well-typed according to Lean4 rules.\\
2.The correction should maintain the original mathematical meaning that the user was likely trying to express in the statement.\\
3.The corrected Lean 4 code must end with ':= by sorry'.\\

For your response, please follow this example format:\\
**Modification Analysis**\\
\verb|```|analysis\\
Your detailed analysis\\
\verb|```|\\
**Corrected Lean4 Code**\\
\verb|```|lean4\\
Your corrected Lean4 code\\
\verb|```|\\

Now, please provide your formal answer:
\end{modify_prompt}
\newenvironment{original_statement_AST}{%
  \begin{tcolorbox}[colback=brown!8!white,colframe=brown!10!black,breakable,title=Original Lean 4 Statement]
}{%
  \end{tcolorbox}
}

\newenvironment{evolved_statement_AST}{%
  \begin{tcolorbox}[colback=brown!8!white,colframe=brown!10!black,title=Evolved Statement]
}{%
  \end{tcolorbox}
}

\subsection{Case Study FOR EvolAST}
\label{appendix:details_for_AST}
We select an example of EvolAST from evolved dataset. The original Lean 4 statement is as follows:
\begin{original_statement_AST}
\begin{minted}[escapeinside=||, texcl=true]{lean}
theorem evolved_thm |$(x, y : \mathbb{R})$| (h_0 : x * y = 4) (h_1 : x > y) (h_2 : x|\textasciicircum|3 - y|\textasciicircum|3 = 3555) : x|\textasciicircum|2 + y|\textasciicircum|2 = 233 := by sorry
\end{minted}
\end{original_statement_AST}

The evolved Lean 4 statement is as follows:

\begin{evolved_statement_AST}
    
Evolved Formal Statement:
\vspace{-10pt} 
    \begin{minted}[escapeinside=||, texcl=true]{lean}
theorem evolved_thm_auged |$(x, y : \mathbb{R})$| (h_1 : y < x) (h_2 : 3555 = x|\textasciicircum|3 - y|\textasciicircum|3) (h_0 : 4 = y * x) :  233 = y|\textasciicircum|2 + x|\textasciicircum|2 := by sorry
\end{minted}
\end{evolved_statement_AST}
\subsection{Use of Large Language Models}
We utilized Large Language Models (LLMs) solely to refine the language and improve the clarity of this manuscript.

\end{document}